\documentclass[letterpaper, 10 pt, conference]{ieeeconf}  

\IEEEoverridecommandlockouts                              

\usepackage{graphics} 
\usepackage{graphicx} 
\usepackage{amsmath} 
\usepackage[noend]{algpseudocode}
\usepackage{algorithmicx,algorithm}
\usepackage{cite}
\usepackage{multirow}
\usepackage{subfigure}
\usepackage{hyperref}

\title{\LARGE \bf
A Multi-task Convolutional Neural Network for Autonomous Robotic Grasping in Object Stacking Scenes
}

\author{Hanbo Zhang, Xuguang Lan, Site Bai, Lipeng Wan, Chenjie Yang, and Nanning Zheng
\thanks{Hanbo Zhang and Xuguang Lan are with the Institute of Artificial Intelligence and Robotics, the National Engineering Laboratory for Visual Information Processing and Applications, School of Electronic and Information Engineering,
        Xi'an Jiaotong University, No.28 Xianning Road, Xi'an, Shaanxi, China.
        {\tt\small zhanghanbo163@stu.xjtu.edu.cn, xglan@mail.xjtu.edu.cn}}%
}

\begin{document}

\maketitle
\thispagestyle{empty}
\pagestyle{empty}

\begin{abstract}

Autonomous robotic grasping plays an important role in intelligent robotics. However, how to help the robot grasp specific objects in object stacking scenes is still an open problem, because there are two main challenges for autonomous robots: (1)it is a comprehensive task to know what and how to grasp; (2)it is hard to deal with the situations in which the target is hidden or covered by other objects. In this paper, we propose a multi-task convolutional neural network for autonomous robotic grasping, which can help the robot find the target, make the plan for grasping and finally grasp the target step by step in object stacking scenes. We integrate vision-based robotic grasping detection and visual manipulation relationship reasoning in one single deep network and build the autonomous robotic grasping system. Experimental results demonstrate that with our model, Baxter robot can autonomously grasp the target with a success rate of 90.6\%, 71.9\% and 59.4\% in object cluttered scenes, familiar stacking scenes and complex stacking scenes respectively.

\end{abstract}

\section{Introduction}

\begin{figure}[t] 
 \center{\includegraphics[scale=0.1]{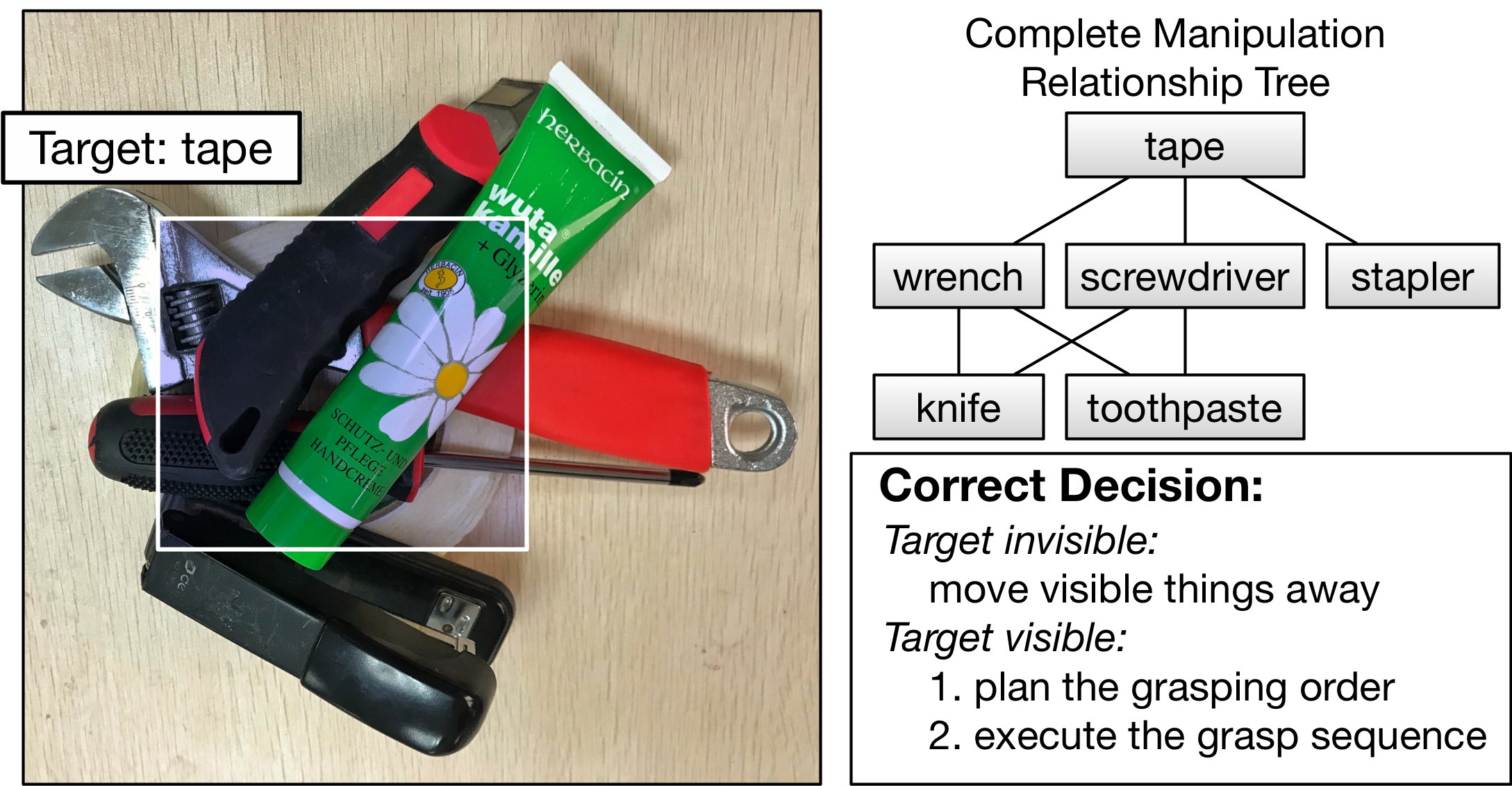}}        
 \caption{Grasping task in a complex scene. The target is the tape, which is placed under several things and nearly invisible. Complete manipulation relationship tree indicates the correct grasping order, and grasping in this order will avoid damages to other objects. The correct decision made by the human being should be: if the target is visible, the correct grasping plan should be made and a sequence of grasps should be executed to get the target while if the target is invisible, the visible things should be moved away in a correct order to find the target.}
 \label{intro}
 \end{figure}

In the research of intelligent robotics, autonomous robotic grasping is a very challenging task \cite{graspsurvey}. For use in daily life scenes, autonomous robotic grasping should satisfy the following conditions:
\begin{itemize}
\item Grasping should be robust and efficient.
\item The desired object can be grasped in a multi-object scene without potential damages to other objects.
\item The correct decision can be made when the target is not visible or covered by other things.
\end{itemize}
For human beings, grasping can be done naturally with high efficiency even if the target is unseen or grotesque. However, robotic grasping involves many difficult steps including perception, planning and control. Moreover, for complex scenes ($e.g.$ the target is occluded or covered by other objects), robots also need certain reasoning ability to grasp the target orderly. For example, as shown in Fig. \ref{intro}, in order to prevent potential damages to other objects, the robot have to plan the grasping order through reasoning, perform multiple grasps in sequence to complete the task and finally get the target. These difficulties make autonomous robotic grasping more challenging in complex scenes.

Therefore, in this paper, we propose a new vision-based multi-task convolutional neural network (CNN) to solve the mentioned problems for autonomous robotic grasping, which can help the robot complete grasping task in complex scenes ($e.g.$ grasp the occluded or covered target). To achieve this, three functions should be implemented including grasping the desired object in multi-object scenes, reasoning the correct order for grasping and executing grasp sequence to get the target. To help the robot grasp the desired object in multi-object scenes, we design the {\bf Perception} part of our network, which can simultaneously detect objects and their grasp candidates. The grasps are detected in the area of each object instead of the whole scene. In order to deal with situations in which the target is hidden or covered by other objects, we design the {\bf Reasoning} part to get visual manipulation relationships between objects and enable the robot to reason the correct order for grasping, preventing potential damages to other objects. For transferring network outputs to configurations of grasping execution, we design {\bf Grasping} part of our network. For the perception and reasoning process, RGB images are taken as input of the neural network, while for execution of grasping, depth information is needed for approaching vector computation and coordinate transformation.
 
Though there are some previous works that try to complete grasping in dense clutter\cite{guo2016object, dexnet3, handeyegrasp,gualtieri2016high}, as we know, our proposed algorithm is the first to combine perception, reasoning, and grasp planning simultaneously by using one neural network, and attempts to realize autonomous robotic grasp in complex scenarios. To evaluate our proposed algorithm, we validate the performance of our model in VMRD dataset\cite{zhang2018visual}. For robotic experiments, Baxter robot is used as the executor to complete grasping tasks, in which the robot is required to find the target, make the plan for grasping and grasp the target step by step.

The rest of this paper is organized as following: Related work is reviewed in Section II; Our proposed algorithm is detailed in Section III; Experimental results including validation on VMRD dataset and robotic experiments are shown in Section IV, and finally the conclusion and discussion are described in Section V.
 
\section {Related Work}

\subsection{Robotic Grasp Detection}

As the development of deep learning, robotic grasp detection based on convolutional neural network (CNN) achieves state-of-the-art performance on several datasets such as Cornell dataset\cite{lenzgrasp, redmongrasp, resnetgrasp, guo2017hybrid, chu2018grasp, zhou2018grasp} and CMU grasp dataset\cite{pinto2016grasp}. They are suitable for grasp detection in single-object scenes. There are some works proposed for grasping in dense clutter\cite{guo2016object, dexnet3, handeyegrasp,gualtieri2016high}. A deep network is used to simultaneously detect the most exposed object and its best grasp by Guo et al.\cite{guo2016object}, which is trained on a fruit dataset including 352 RGB images. However, their model can only output the grasp affiliated to the most exposed object without perception and understanding of the overall environment and reasoning of the relationship between objects, which limits the use of the algorithm. Algorithms proposed in \cite{dexnet3}, \cite{handeyegrasp} and \cite{gualtieri2016high} only focus on the detection of grasps in scenes where objects are densely cluttered, rather than what the grasped objects are. Therefore, the existing algorithms detect grasps on features of the whole image, and can only be used to grasp an unspecified object instead of a pointed one in stacking scenes.
 
\subsection{Visual Manipulation Relationship Reasoning}

Recent works prove that CNNs achieve advanced performance on visual relationship reasoning\cite{vrd, rlvrd, vrd2017dai}. Different from visual relationship, visual manipulation relationship\cite{zhang2018visual} is proposed to solve the problem of grasping order in object stacking scenes with consideration of the safety and stability of objects. However, when this algorithm is directly combined with the grasp detection network to solve grasping problem in object stacking scenes, there are two main difficulties: 1) it is difficult to correctly match the detected grasps and the detected objects in object stacking scenes; 2) the cascade structure causes a lot of redundant calculations ($e.g.$ the extraction of scene features), which makes the speed slow.\\

Therefore, in this paper, we propose a new CNN architecture to combine object detection, grasp detection and visual manipulation relationship reasoning and build a robotic autonomous grasping system. Different from previous works, the grasps are detected on the object features instead of the whole scene. Visual manipulation relationships are applied to decide which object should be grasped first. The proposed network can help our robot grasp the target following the correct grasping order in complex scenes.
 
\section{Task Definition}

In this paper, we focus on grasping task in scenes where the target and several other objects are cluttered or piled up, which means there can be occlusions and overlaps between objects, or the target is hidden under other objects and cannot be observed by the robot. Therefore, we set up an environment with several different objects each time. In each experiment, we test whether the robot can find out and grasp the specific target. The target of the task is input manually. The desired robot behavior is that the final target can be grasped step by step following the correct manipulation relationship predicted by the proposed neural network.

  \begin{figure}[t] 
 \center{\includegraphics[scale=1]{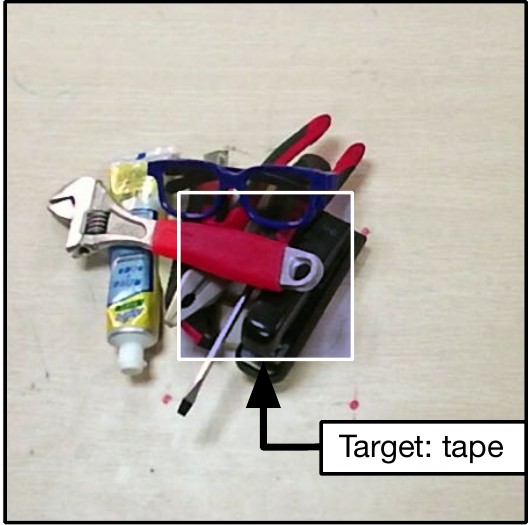}}        
 \caption{Our task is to grasp the target in complex scenes. The target is covered by several other objects and almost invisible. The robot need to find the target, plan the grasping order and execute the grasp sequence to get the target.}
 \label{taskdef}
 \end{figure}

In detail, we focus on grasping task in realistic and challenging scenes as following: each experimental scene includes 6-9 objects, where objects are piled up and there are severe occlusions and overlaps. In the beginning of each experiment, the target is difficult to detect in most cases. Following this setting, it can test whether the robot can make correct decisions to find the target and successfully grasp it.

\section{Proposed Approach}

\subsection{Architecture}

\begin{figure*}[t] 
 \center{\includegraphics[width=\textwidth]{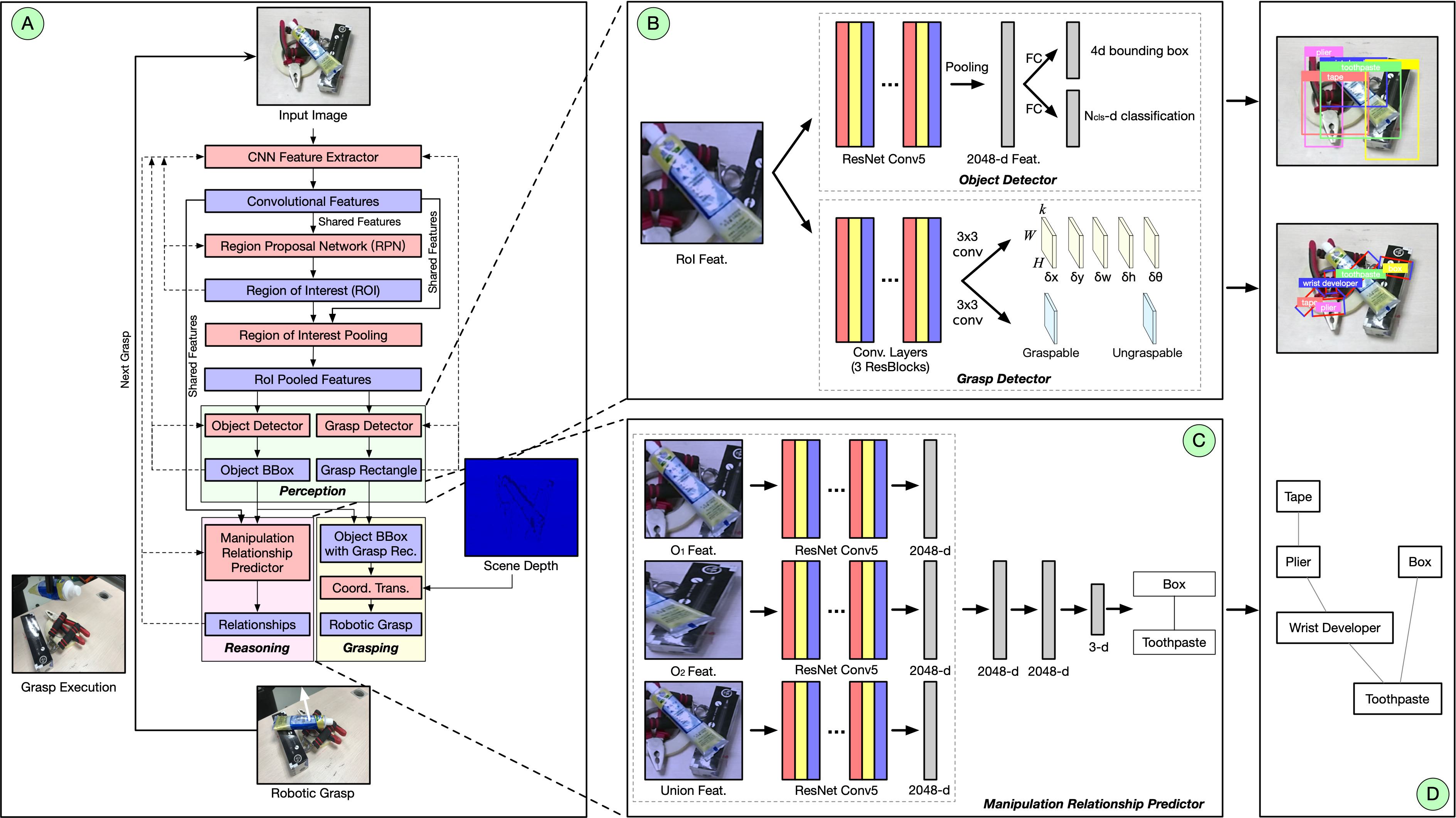}}        
 \caption{Architecture of our proposed approach. The input is RGB images of working scenes. The solid arrows indicate forward-propagation while the dotted arrows indicate backward-propagation. In each iteration, the neural network produces one robotic grasp configuration and the robot moves one object. The iteration will not be terminated until the desired target is grasped. (a): Network architecture; (b): Perception part with object detector and grasp detector; (c): Reasoning part with visual manipulation relationship predictor; (d): Expected results.}
 \label{arch}
 \end{figure*}
 
The proposed architecture of our approach is shown in Fig. \ref{arch}. The input of our network is RGB images of working scenes. First, we use CNN ($e.g.$ ResNet-101\cite{resnet}) to extract image features. As shown in Fig. \ref{arch}, convolutional features are shared among Region proposal network (RPN, \cite{fasterrcnn}), object detection, grasp detection and visual manipulation relationship reasoning. The shared feature extractor used in our work is ResNet-101 layers before conv4 including 30 ResBlocks. Therefore, the stride of shared features is 16. 

RPN follows the feature extractor to output regions of interest (ROI). RPN includes three $3\times3$ convolutional layers: an intermediate convolutional layer, a ROI regressor and a ROI classifier. The ROI regressor and classifier are both cascaded after the intermediate convolutional layer to output locations of ROIs and the probability of each ROI being a candidate of object bounding boxes.

The mainbody of our approach includes 3 parts: Perception, Reasoning and Grasping. ``Perception" is used to obtain the detection results of object and grasp with the affiliation between them. ``Reasoning" takes object bounding boxes output by ``Perception" and image features as input to predict the manipulation relationship between each pair of objects. ``Grasping" uses perception results to transform grasp rectangles into robotic grasp configurations to be executed by the robot. Each detection produces one robotic grasp configuration, and the iteration is terminated when the desired target is grasped.

\subsection{Perception}

In ``Perception" part, the network simultaneously detects objects and  their grasps. The convolutional features and ROIs output by RPN are first fed into ROI pooling layer, where the features are cropped by ROIs and pooled using adaptive pooling into the same size $W\times H$ (in our work, the size is $7\times7$). The purpose of ROI pooling is to enable the corresponding features of all ROIs to form a batch for network training.

\subsubsection {Object Detector} Object detector takes a mini-batch of ROI pooled features as input. As in \cite{resnet}, a ResNet' conv5 layer including 9 convolutional layers is adopted as the header for final regression and classification taking ROI pooled features as input. The header's output is then averaged on each feature map. The regressor and classifier are both fully connected layers with 2048-d input and no hidden layer, outputting locations of refined object bounding boxes and classification results respectively.

\subsubsection {Grasp Detector} Grasp detector also takes ROI pooled features as input to detect grasps on each ROI. Each ROI is firstly divided into $W\times H$ grid cells. Each grid cell corresponds to one pixel on ROI pooled feature maps. Inspired by our previous work\cite{zhou2018grasp}, the grasp detector outputs $k$ (in this paper, $k=4$) grasp candidates on each grid cell with oriented anchor boxes as priors. Different oriented anchor size is explored during experiments including $12\times12$ and $24\times24$ pixels. A header including 3 ResBlocks cascades after ROI pooling in order to enlarge receptive field of features used for grasp detection. The reason is that a large receptive field can prevent grasp detector from being confused by grasps that belongs to different ROIs. Then similar to \cite{zhou2018grasp}, the grasp regressor and grasp classifier follow the grasp header and output $5k$ and $2k$ values for grasp rectangles and graspable confidence scores respectively. Therefore, the output for each grasp candidate is a 7-dimension vector, 5 for the location of the grasp $(\delta x_g,\delta y_g,\delta w_g,\delta h_g,\delta \theta_g)$ and 2 for graspable and ungraspable confidence scores $(c_g,c_{ug})$.

Therefore, the output of ``Perception" part for each object contains two parts: object detection result $O$ and grasp detection result $G$. $O$ is a 5-dimension vector $(x_{min},y_{min},x_{max},y_{max},cls) $ representing the location and category of an object and $G$ is a 5-dimension vector $(x_g,y_g,w_g,h_g,\theta_g)$ representing the best grasp. $(x_g,y_g,w_g,h_g,\theta_g)$ is computed by Eq. (1):
\begin{equation}
\begin{split}
x_g &= \delta x_g \times w_{a} + x_{a}\\ 
y_g &= \delta y_g \times h_{a} + y_{a}\\
w_g &= exp(\delta w_g) \times w_a\\
h_g &= exp(\delta h_g) \times h_a\\
\theta_g&=\delta \theta_g \times (90/k) + \theta_a
\end{split}
\label{encode}
\end{equation}
where $(x_a,y_a,w_a,h_a,\theta_a)$ is the corresponding oriented anchor.

\begin{figure}[t] 
 \center{\includegraphics[width=0.45\textwidth]{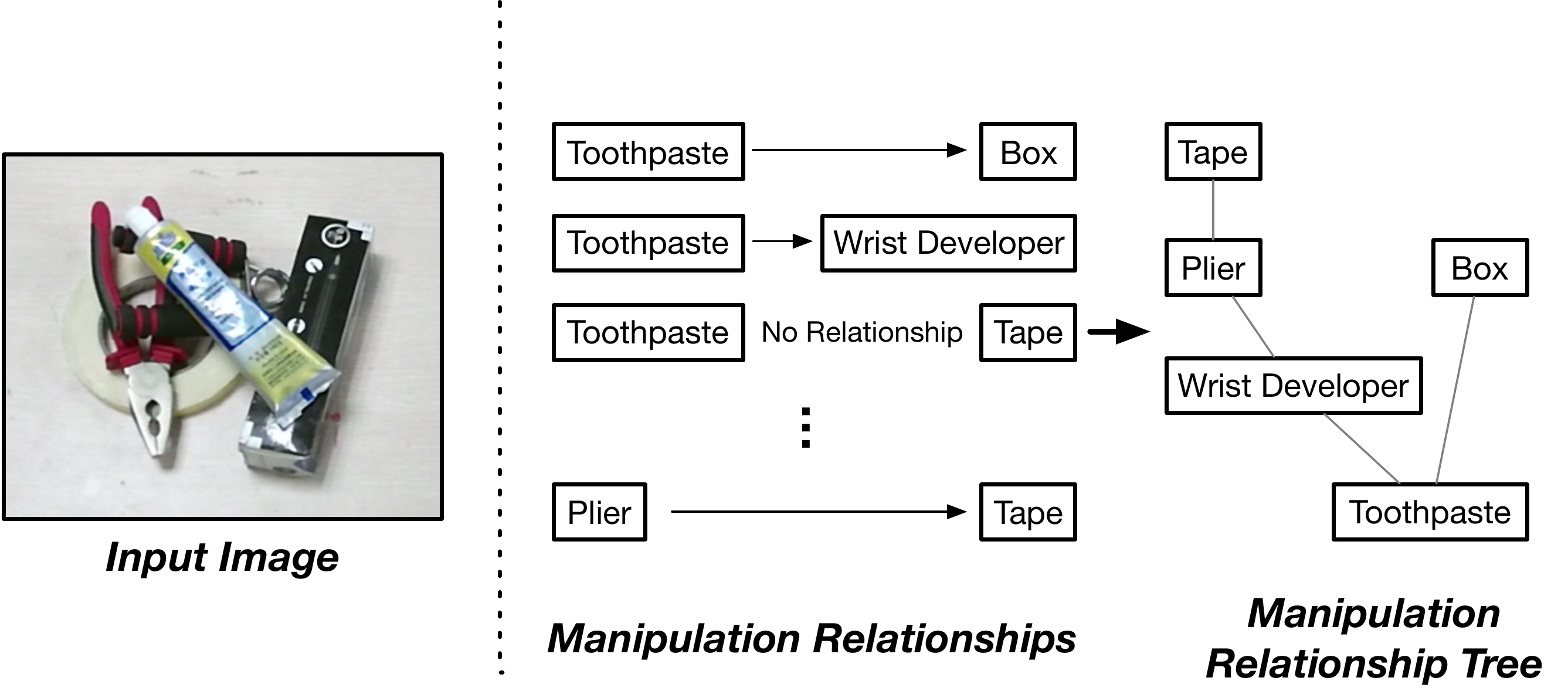}}        
 \caption{When all manipulation relationships are obtained, the manipulation relationship tree can be built combining all manipulation relationships. The leaf nodes represent the objects that should be grasped first.}
 \label{maniputree}
 \end{figure}

\subsection{Reasoning}

Inspired by our previous work\cite{zhang2018visual}, we combine visual manipulation relationship reasoning in our network to help robot reason the grasping order without potential damages to other objects. 

\emph{Manipulation Relationship Predictor:} To predict manipulation relationships of object pairs, we adopt Object Pairing Pooling Layer (OP$^2$L) to obtain features of object pairs. As shown in Fig. \ref{arch}.C, the input of manipulation relationship predictor is features of object pairs. The features of each object pair $(O_1,O_2)$ include the features of $O_1$, $O_2$ and the union bounding box. Similar to object detector and grasp detector, the features of each object are also adaptively pooled into the same size of $W\times H$. The difference is that the convolutional features are cropped by object bounding boxes instead of ROIs. Note that $(O_1,O_2)$ is different from $(O_2,O_1)$ because manipulation relationship does not conform to the exchange law. If there are $n$ detected objects, the number of object pairs will be $n\times(n-1)$, and there will be $n\times(n-1)$ manipulation relationships predicted. In manipulation relationship predictor, the features of the two objects and the union bounding box are first passed through several convolutional layers respectively (in this work, ResNet Conv5 layers are applied), and finally manipulation relationships are classified by a fully connected network containing two 2048-d hidden layers.

After getting all manipulation relationships, we can build a \emph {manipulation relationship tree} to describe the correct grasping order in the whole scene as shown in Fig. \ref{maniputree}. Leaf nodes of the manipulation relationship tree should be grasped before the other nodes. Therefore, it is worth noting that the most important part of the manipulation relationship tree is the leaf nodes. In other words, if we can make sure that the leaf nodes are detected correctly in each step, the grasping order will be correct regardless of the other nodes.

\subsection{Grasping}
 
``Grasping" part is used to complete inference on outputs of the network. In other words, the input of this part is object and grasp detection results, and the output is the corresponding robotic configuration to grasp each object. Note that there is no trainable weights in ``Grasping" part.

\subsubsection{Grasp Selection} As described above, the grasp detector outputs a large set of grasp candidates for each ROI. Therefore, the best grasp candidate should be selected first for each object. According to \cite{chu2018grasp}, there are two methods to find the best grasp: (1) choose the grasp with highest graspable score; (2) choose the one closest to the object center in Top-$N$ candidates. The second one is proved to be a better way in \cite{chu2018grasp}, which is used to get the grasp of each object in our paper. In our experiments, $N$ is set to 3.

\subsubsection{Coordinate Transformation} The purpose of the coordinate transformation is to map the detected grasps in the image to the approaching vector and grasp point in the robot coordinate system. In this paper, an affine transformation is used approximately for this mapping. The affine transformation is obtained through four reference points with their coordinates in the image and robot coordinate system. The grasp point is defined as the point in grasp rectangle with minimum depth while the approaching vector is the average surface normal around the grasp point. The grasp point and approaching vector will be mapped into robot coordinate system for location of the robot gripper in grasp execution.

So far, the robot knows which object should be firstly grasped by ``Reasoning", where to grasp by ``Perception" and how to grasp by ``Grasping". By following these steps, objects will be grasped one by one until the target is obtained.
 
\subsection{Training}

Our networks are trained end-to-end with one multi-task loss function. The loss function includes three parts: object detection loss $L_O$, grasp detection loss $L_G$ and visual manipulation relationship reasoning loss $L_R$, where $L_O$ is same as \cite{fasterrcnn} and $L_R$ is a multi-class Negative Log-Likelihood classification loss as shown in Eq. (\ref{lr}):
\begin{equation}
L_R = -\sum_{(O_i,O_j)} log (p^{(O_i,O_j)}_r)
\label{lr}
\end{equation}
where $r\in\{0,1,2\}$ is the ground truth relationship between $O_i$ and $O_j$ and $p^{(O_i,O_j)}_r$ is the predicted probability that the object pair $(O_i,O_j)$ has the relationship $r$.

$L_G$ is designed to simultaneously minimize the grasp rectangle regression loss and classification loss. As described above, each oriented anchor $(x_a,y_a,w_a,h_a,\theta_a)$ will be corresponding to 5-dimension offsets $(\delta x_g, \delta y_g, \delta w_g,\delta h_g, \delta \theta_g)$ and 2-dimension of confidence scores $(c_g, c_{ug})$. Therefore, grasp detection loss is defined as following:
\begin{equation}
L_G = L_{Greg} + \alpha L_{Gcls}
\label{lg}
\end{equation}
with
\begin{equation}
L_{Greg} = \sum_{i\in Positive}^P  \sum_{m\in \{x,y,w,h,\theta\}}^{}smoothL1(\delta m_g^{(i)}-\delta m_{gt}^{(i)})
\end{equation}
\begin{equation}
L_{Gcls} = -\sum_{i\in Positive}^P log(c^{(i)}_{g}) - \sum_{i\in Negative}^{3P} log(c^{(i)}_{ug})
\end{equation}
where $\delta m_{gt},m\in \{x,y,w,h,\theta\}$ represents the ground truth offset and $P$ is the number of oriented anchors that match at least one ground truth. If an oriented anchor is not matched to any ground truth, it will be treated as a negative sample. We select the top-$3P$ negative samples for classification training. According to experience, $\alpha$ is 1 in this paper. The whole loss for our network is defined as follow:
\begin{equation}
\begin{split}
loss = L_{O} + \lambda L_{G} + \beta L_R\\
\end{split}
\label{lwhole}
\end{equation}
where $\lambda$ and $\alpha$ are all set to 1 in our experiments.

\section{Experiment}
 
\begin{figure}[t] 
 \center{\includegraphics[width=0.48\textwidth]{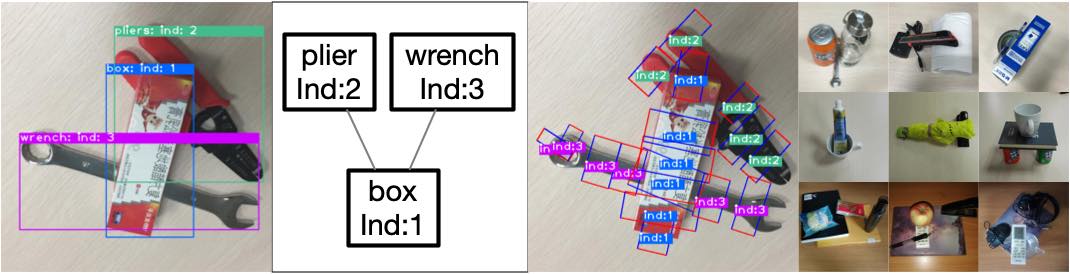}}        
 \caption{Examples of VMRD dataset with grasps. Each grasp is labeled with an index that indicates the owner of the grasp.}
 \label{dataset}
 \end{figure}

\subsection{Dataset}
 
Dataset used to train our models is VMRD dataset with grasps\cite{zhang2018visual}. There are 4683 images with grasps in total, which are divided into training set and testing set with 4233 and 450 images respectively. With consideration of the affiliation between objects and grasps, we define the object to which certain grasps are affiliated as their owner. In VMRD, more than 100k grasps are labeled on these images with object indices that indicate the owner of these grasps. In each image, there are 2-5 objects stacked together with overlaps and occlusions. Manipulation relationships are also labeled as manipulation relationship tree, where the leaf nodes should be grasped before the others. One example is shown in Fig. \ref{dataset}.

During training, we take advantages of online data augmentation including random brightness, random hue, random contrast, color space conversion, vertical rotation and horizontal flip, That is to say, in each iteration, images feed into network are different from all the previous, which can prevent overfitting and enhance the generalization ability of our model in different workspace. Note that the image preprocessing does not change the image size including reshape and crop. Before the images are fed into the network, we approximately subtract 144 on each pixel to mean-center all the inputs.

\subsection{Training Details}

Our networks are trained end-to-end on GTX 1080Ti with 11GB memory using PyTorch as the deep learning platform. Learning rate is set to 0.001 according to experience, which is divided by 10 every 20000 iterations. Limited by the memory of GPU, size of mini-batch in this paper is set to 2. We use SGD as the optimizer and set momentum to 0.9. The other settings are same with \cite{fasterrcnn}.
 
\subsection{Metrics}

Note that even though our multi-task network is trained and tested end-to-end, we evaluate the performance of each task separately. Therefore, to evaluate our proposed network, metrics of our experiments also include 3 parts: 1) Perception; 2) Reasoning; 3) Grasping. 

Perception metrics are used to evaluate the performance of the perception output, including object and grasp detection. In this part, we combine the results of object and grasp detection to see how well our network performs on the test set of VMRD. Similarly, reasoning metrics are used to test the performance of the reasoning output on the test set of VMRD. This part will take our previous work\cite{zhang2018visual} as the baseline. Grasping metrics are used to evaluate how well the proposed network performs in real-world experiments.

 \begin{table}[b]
\caption{Validation Results of Perception}
\label{perception}
\begin{center}
\begin{tabular}{l|c|c}
\hline
\multirow{2}{*}{\bf Setting} & {\bf mAP with grasp} &{\bf Speed}\\
& {\bf  (\%)} & {\bf  (FPS)}\\
\hline 
$12\times12$ Anchor, k=4 & 68.0 & \multirow{2}{*}{7.6}\\
$24\times24$ Anchor, k=4 &  65.0 \\
\hline
 $12\times12$ Anchor, k=4, Hi-res & 69.1 & \multirow{2}{*}{6.5}\\
$24\times24$ Anchor, k=4, Hi-res &  {\bf70.5} & \\
\hline
\end{tabular}
\end{center}
\end{table}
 
\subsection{Result}

\begin{table*}[t]
\caption{Validation Results of Reasoning}
\label{reasoning}
\begin{center}
\begin{tabular}{l|c|c|c|c|c|c|c|c}
\hline
\multirow{3}{*}{{\bf Algorithm}} & \multirow{3}{*}{{\bf Setting}} & \multirow{3}{*}{{\bf Obj. Rec.}} & \multirow{3}{*}{{\bf Obj. Prec.}} &  \multicolumn{5}{c}{{\bf Image Acc.}}\\
\cline{5-9}
& & & & \multirow{2}{*}{\begin{minipage}{0.7cm}\centering Total (\%)\end{minipage}} & \multicolumn{4}{c}{Object Number per Image}  \\
\cline{6-9}
& & & & & \begin{minipage}{0.7cm}\centering2\end{minipage} & \begin{minipage}{0.7cm}\centering3\end{minipage} & \begin{minipage}{0.7cm}\centering4\end{minipage} & \begin{minipage}{0.7cm}\centering5\end{minipage}\\
\hline
VMRN (baseline)\cite{zhang2018visual} & - & 82.3 & 78.0 & 63.1 & - & - & - & - \\
\hline
\multirow{4}{*}{{\bf Ours}}  & $12\times12$ Anchor, k=4 & 85.7& 87.2 & 66.4 & 61/65 & 130/209 & 55/106 & 53/70 \\
& $24\times24$ Anchor, k=4& {\bf86.0} &  {\bf 88.8} & {\bf67.1} & 57/65 & 134/209 & 60/106 & 51/70 \\
& $12\times12$ Anchor, k=4, Hi-res & 85.9 & 85.8 & 65.3 & 61/65 & 132/209 & 51/106 & 49/70 \\
& $24\times24$ Anchor, k=4, Hi-res & 84.7 & 86.5 & 65.1 & 60/65 & 133/209 & 55/106 & 45/70 \\
\hline
\end{tabular}
\end{center}
\end{table*}

   \begin{figure*}[t] 
 \center{\includegraphics[width=0.9\textwidth]{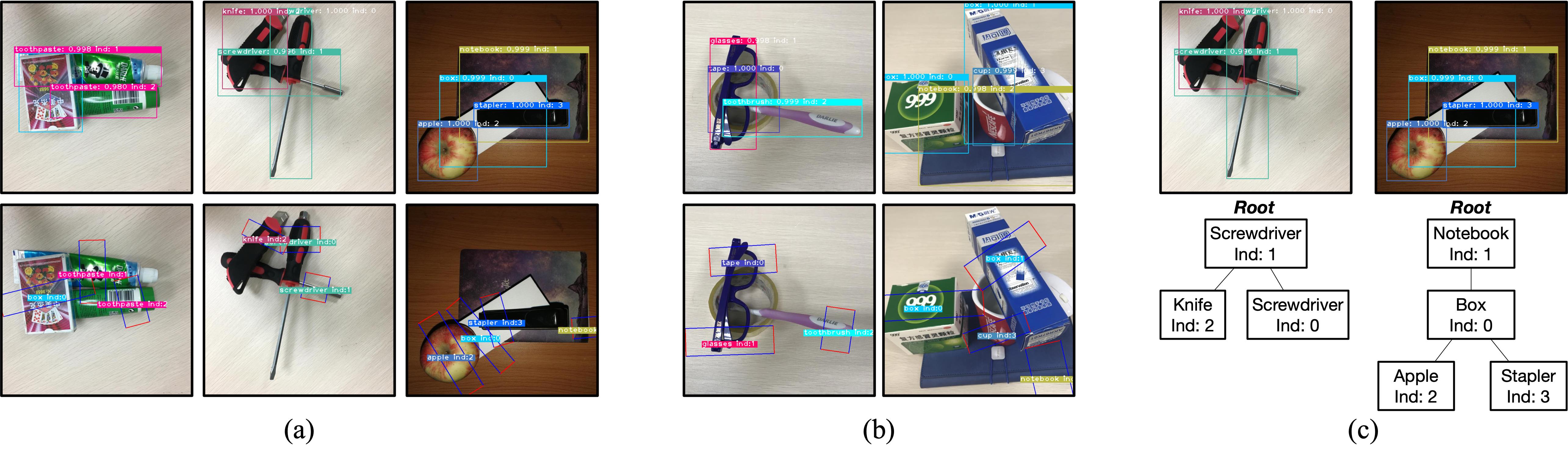}}        
 \caption{Selected results of ``Perception" and ``Reasoning". (a) True positive examples for detection of object and grasp. (b) Images with incorrect detections. In (a) and (b), the top row is object detection results and the second row is grasp detection results. (c) Examples of manipulation relationship reasoning.}
 \label{perceptionexample}
 \end{figure*}
 
\subsubsection{Perception} Perception results are shown in Table \ref{perception}, considering detection results of object and grasp simultaneously. An object is thought as a true positive detection when the bounding box $O$ and the best grasp $G$ are all correctly detected. In detail, the detection $(O,G)$ should satisfy the following conditions:
\begin{itemize}
\item The bounding box $O$ should have an IoU larger than 0.5 with the ground truth and be classified correctly
\item The best grasp $G$ should has a Jaccard Index larger than 0.25 and angle difference less than $30^{\circ}$ with at least one ground truth grasp
\end{itemize}
Formally, a detection $(O,G)$ is a 10-d vector: 
\begin{equation}
(x_{min},y_{min},x_{max},y_{max},cls,x_g,y_g,w_g,h_g,\theta_g)
\end{equation}
which includes 5 more values $(x_g,y_g,w_g,h_g,\theta_g)$ than only object detection, indicating the best grasp of the corresponding object. Therefore, we can use the same measurement \emph{mAP} in object detection to evaluate the performance of our model, with consideration of whether the best grasp is correct or not. In Table \ref{perception}, \emph{mAP with grasp} combines object detection $O$ and grasp detection $G$ together to measure the performance of the perception. \emph{Hi-res} means the input image of the neural network is high-resolution (from 600 pixels to 800 pixels in our experiments).
 
Some examples are shown in Fig. \ref{perceptionexample}. From the true positive examples in Fig. \ref{perceptionexample}(a), we can see that our model can successfully detect objects with their grasps. Failure cases are shown in Fig. \ref{perceptionexample}(b). We can see that the excessive overlap between objects will make it difficult for the model to find the proper grasp for each object. Besides, occlusions and overlaps also make trouble for object detection.
 
\subsubsection{Reasoning}

Reasoning results are shown in Table \ref{reasoning} and Fig. \ref{perceptionexample}(c). Following \cite{zhang2018visual}, \emph{Obj. Rec.} and \emph{Obj. Prec.} are the recall and precision when the object pair is considered as a sample: it is treated as a positive only when two objects are detected  and the relationship between them is reasoned correctly. \emph{Image Acc.} is the image accuracy, in which the image is thought to be a positive only when all objects in it are detected and all relationships are reasoned correctly. 

We can see that our model improves the performance of visual manipulation relationship reasoning compared with our previous work \cite{zhang2018visual}. We assume that the improvements are achieved following these changes: (1) Different from our previous work, in this paper, we use ResNet-101 as the feature extractor, or called ``backbone" instead of ResNet-50 and VGG-16 in \cite{zhang2018visual}; (2) the object detector is Faster-RCNN from \cite{fasterrcnn} instead of SSD in \cite{ssd}; (3) the backbone is updated using multi-task loss function including grasp detection loss.

Two examples in Fig. \ref{perceptionexample}(c) demonstrate the output of visual manipulation relationship reasoning. We can see that our model can efficiently build the manipulation relationship tree for grasping task in object stacking scenes. Note that for scenes where there are more than 1 objects belonging to the same category (like shown in the first example of Fig. \ref{perceptionexample}(c), there are two screwdrivers), our model can also work well by giving each object a unique ID (called ``index" in VMRD) to distinguish them.

\subsubsection{Grasping} 

\begin{figure}[t] 
 \center{\includegraphics[width=0.25\textwidth]{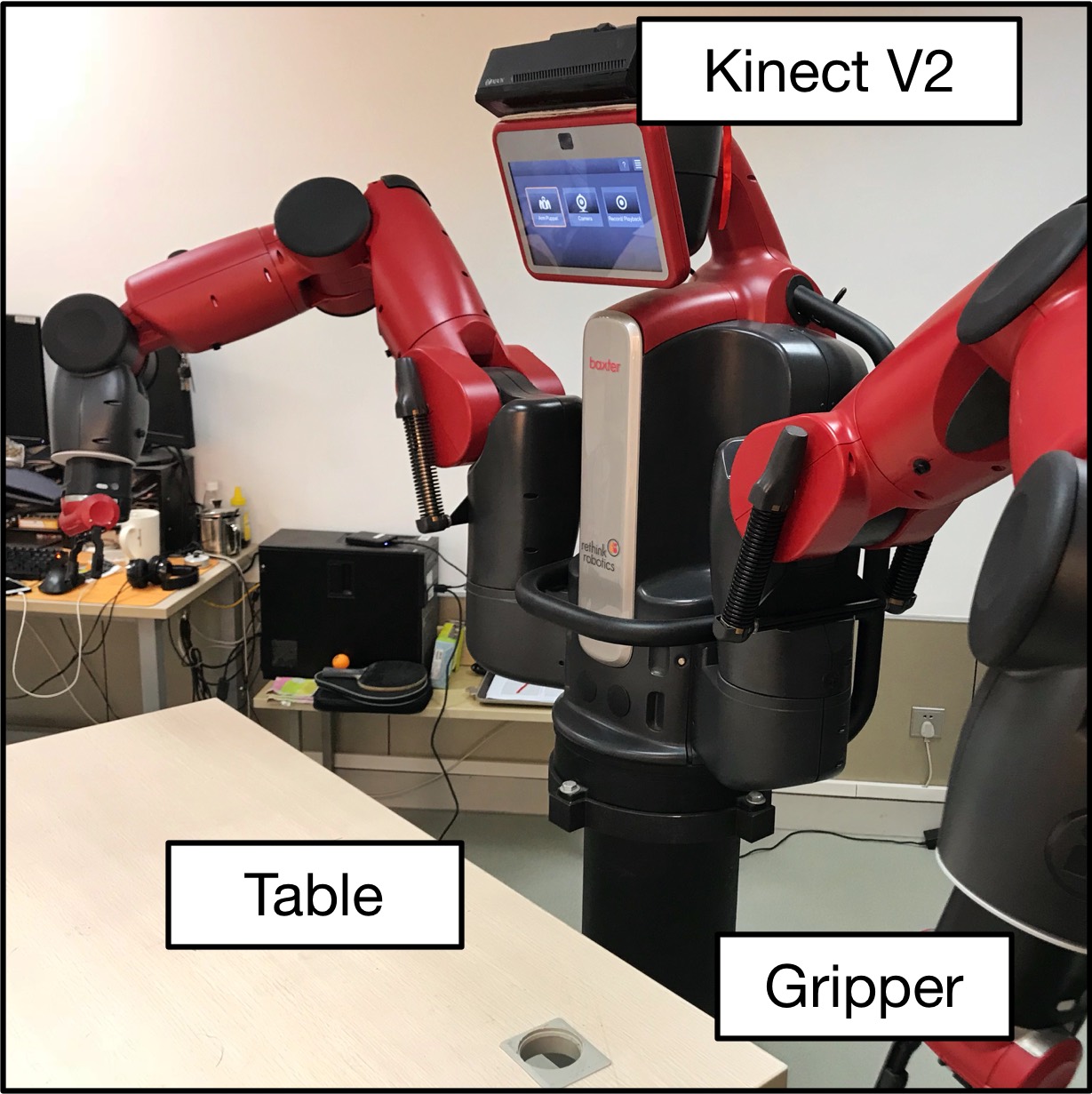}}        
 \caption{Robot environment}
 \label{env}
 \end{figure}

\begin{table}[t]
\caption{Robotic Experimental Results}
\label{robexp}
\begin{center}
\begin{tabular}{l|c|c}
\hline
\multirow{3}{*}{Scene Setting} & \multicolumn{2}{c}{Success Rate of Each Iteration} \\
\cline{2-3}
& Baseline & Ours\\
\hline
Object Cluttered Scenes & 43.8\% (14/32) & {\bf 90.6\% (29/32)} \\
Familiar Stacking Scenes & 28.1\% (9/32) & {\bf 71.9\% (23/32)} \\
Complex Stacking Scenes &  6.3\% (2/32) & {\bf 59.4\% (19/32)} \\
\hline
\end{tabular}
\end{center}
\end{table}
 
  \begin{figure}[t] 
 \center{\includegraphics[width=0.35\textwidth]{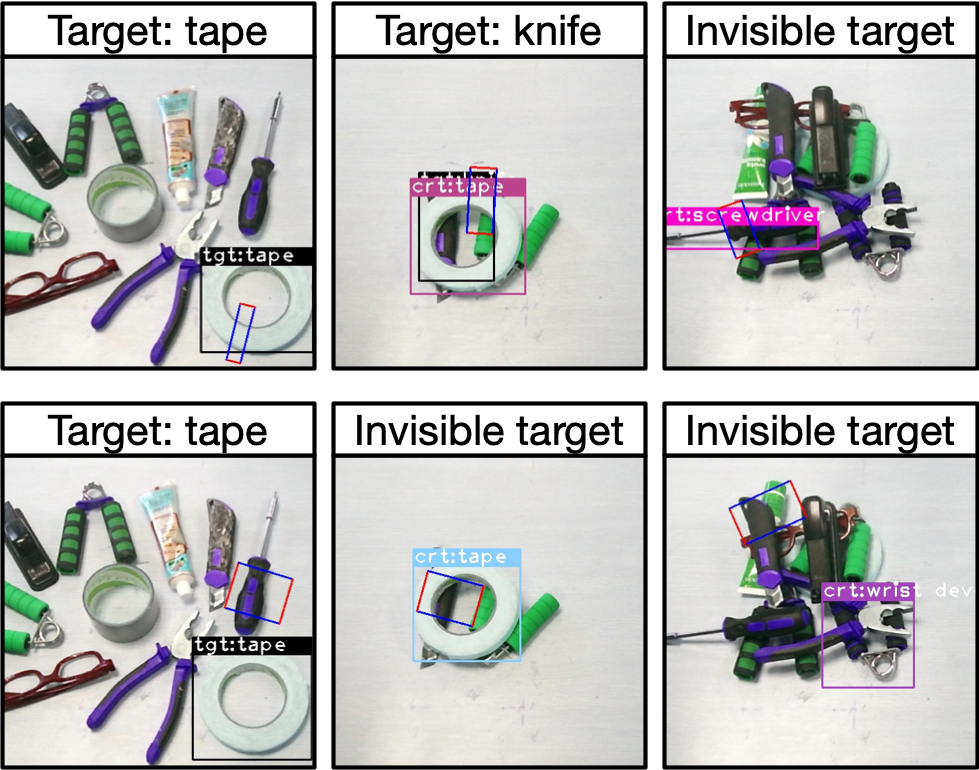}}        
 \caption{Comparison between the baseline and our proposed network. The top row is the results of our network and the second row is the results output by the baseline. Final target is abbreviated by ``tgt" and the immediate object to be grasped is abbreviated by ``crt" (current).}
 \label{compfig}
 \end{figure}

\begin{figure}[t] 
 \center{\includegraphics[width=0.42\textwidth]{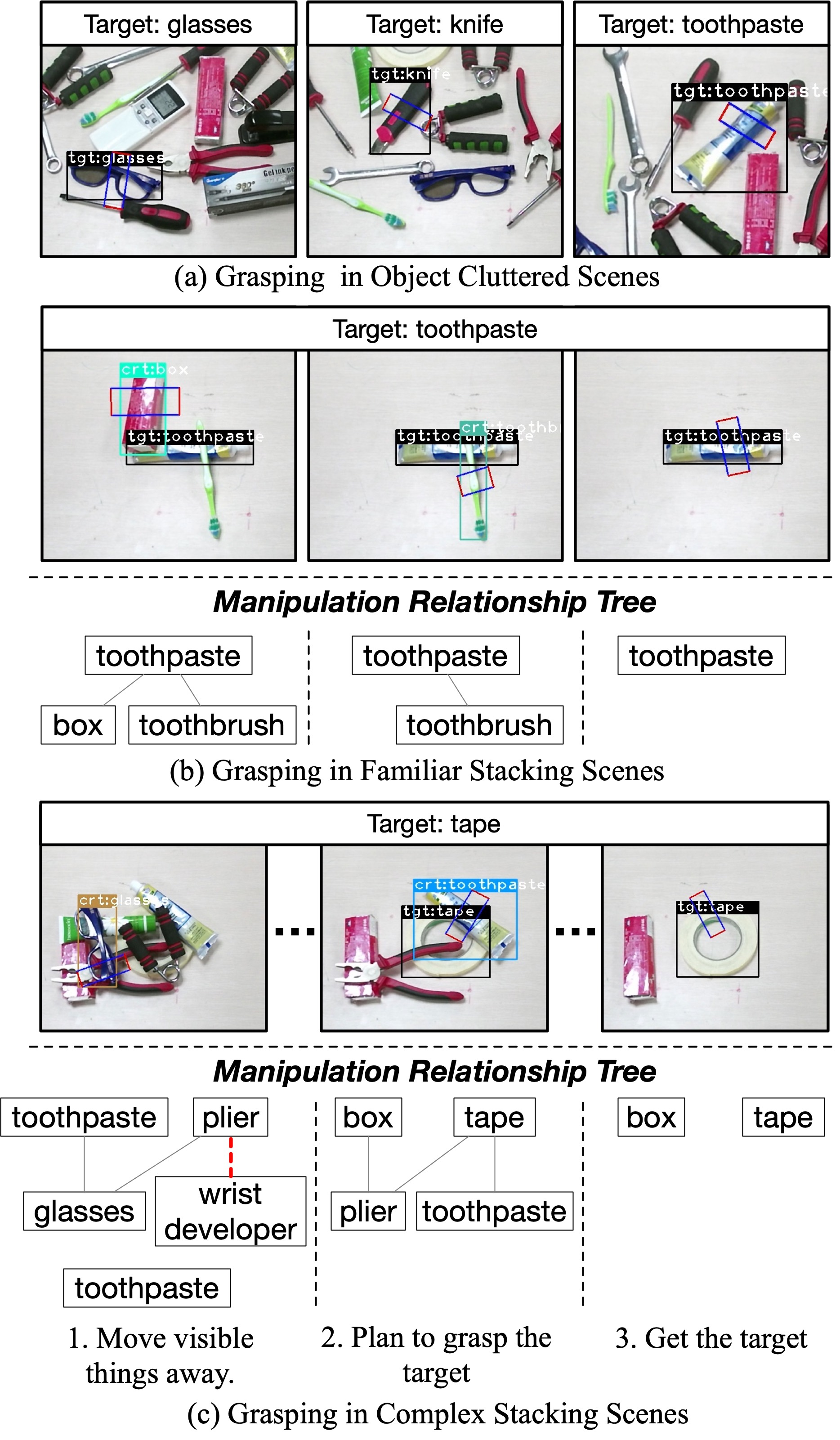}}        
 \caption{Examples of successful grasping under three conditions. Final target is abbreviated by ``tgt" and the immediate object to be grasped is abbreviated by ``crt" (current). The red dotted line represents a detection error: the relationship between the wrist developer and the plier is not detected correctly.}
 \label{robexpexample}
 \end{figure}

\begin{figure}[t] 
 \center{\includegraphics[width=0.45\textwidth]{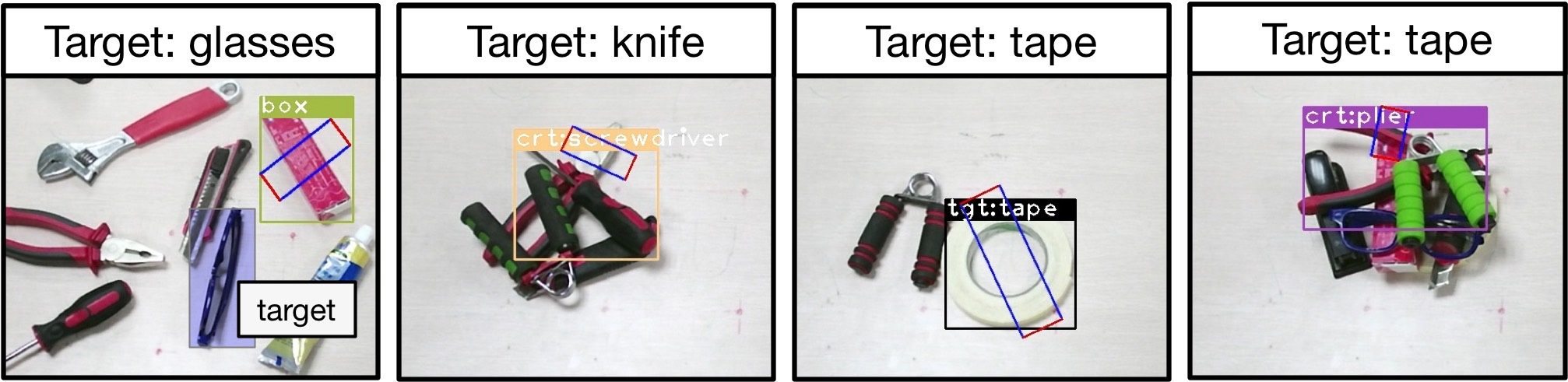}}        
 \caption{Failures during robot experiment.}
 \label{robexpfailures}
 \end{figure}

 
In this part, robotic grasping experimental results are explored using the model with $12\times12$ anchors and Hi-res inputs. Our robotic experiments are conducted with targets being specified in three types of scenes: 
\begin{itemize}
\item \emph{Object Cluttered Scenes} where objects are scattered and the target should be directly grasped by the robot.
\item \emph{Familiar Stacking Scenes} where there are 2-5 objects stacked like in VMRD dataset. The object number in each scene is chosen uniformly from $\{2,3,4,5\}$.
\item \emph{Complex Stacking Scenes} where there are 6-9 objects stacked together. The object number in each scene is chosen uniformly from $\{6,7,8,9\}$.
\end{itemize}
Among them, the most challenging task is grasping in \emph{Complex Stacking Scenes}. A successful experiment for accomplishing the task is defined as: following the correct grasping order, the specific target is grasped successfully.

We use Baxter robot as the executor and Kinect v2 as the camera. For the specific usage of Kinect v2, RGB images are applied in perception and reasoning, while depth information is added for computing the grasp point and approaching vector in robot coordinate system. Robotic environment is shown in Fig. \ref{env}.

Experimental results are shown in Table \ref{robexp}. To demonstrate the advantages of our multi-task network, we cascade state-of-the-art model of VMRN in \cite{zhang2018visual} and fully convolutional grasp detection network (FCGN) in \cite{zhou2018grasp} as the baseline. Scenes used to test the baseline and our proposed multi-task network are designed as the same. From the table, we can see that our model works well under all three conditions and achieves a success rate of 90.6\%, 71.9\% and 59.4\% respectively, outperforming the baseline by 46.8\%, 43.8\% and 53.1\%. The success rate will decrease with the growth of scene complexity. Some comparisons between the baseline and our proposed network are shown in Fig. \ref{compfig}. Failures of the baseline are mainly caused by: 1) FCGN is designed for grasp detection in single-object scenes and trained on Cornell Grasp Dataset, hence the generalization ability is not satisfactory in multi-object scenes and it cannot detect grasps for all objects; 2) the affiliations between grasps and their owner are likely to be incorrect.

Selected examples of our robotic experiments are shown in Fig. \ref{robexpexample} and some failures are demonstrated in Fig. \ref{robexpfailures}. In the first picture, target detection failure is caused by the low confidence score of the glasses. As described in the previous section, if the robot cannot see the target, it will move visible things away to find it. Therefore, the box is selected. The second picture demonstrates incorrect grasp detection, which is caused by the confusion of grasps belonging to two different objects due to excessive object overlap. The third picture shows an oversized grasp for our robot. The incorrect order is come across occasionally like shown in the last picture.

\section{Conclusions}

In this paper, we propose a multi-task deep network for autonomous robotic grasping in complex scenes, which can help robot find the target, make the plan for grasping and finally grasp the target step by step in object stacking scenes. Experiments demonstrate that with our model, Baxter robot can autonomously grasp the target with a success rate of 90.6\%, 71.9\% and 59.4\% in object cluttered scenes, familiar stacking scenes and complex stacking scenes respectively. In future work, we will try to overcome object and grasp detection difficulty in excessive overlap scenes for better performance of our model.

\addtolength{\textheight}{-5 cm}   



\section*{ACKNOWLEDGMENT}

This work was supported in part by the key project of Trico-Robot plan of NSFC under grant No.91748208, National Science and Technology Major Project  No. 2018ZX01028-101, key project of Shaanxi province No.2018ZDCXLGY0607, and NSFC No.61573268.

\bibliographystyle{unsrt}
\bibliography{zhbbib}

\end{document}